\pdfoutput=1

\documentclass[11pt]{article}

\usepackage[]{acl}

\usepackage{times}
\usepackage{latexsym}
\usepackage{graphicx}
\usepackage{mathtools}

\usepackage[T1]{fontenc}

\usepackage[utf8]{inputenc}

\usepackage{microtype}

\title{Gender Bias in BERT - Measuring and Analysing Biases through Sentiment Rating in a Realistic Downstream Classification Task}

\author{Sophie F. Jentzsch \\
  Institute for Software Technology  \\
  German Aerospace Center (DLR) \\
  Cologne, Germany \\
  \texttt{Sophie.Jentzsch@DLR.de} \\\And
  Cigdem Turan \\
  Dept. of Computer Science \\
  TU Darmstadt \\
  Darmstadt, Germany \\
  \texttt{cigdem.turan@cs.tu-darmstadt.de} \\}

\begin{document}
\maketitle
\begin{abstract}
Pretrained language models are publicly available and constantly finetuned for various real-life applications. As they become capable of grasping complex contextual information, harmful biases are likely increasingly intertwined with those models. This paper analyses gender bias in BERT models with two main contributions: First, a novel bias measure is introduced, defining biases as the difference in sentiment valuation of female and male sample versions. Second, we comprehensively analyse BERT's biases on the example of a realistic IMDB movie classifier. By systematically varying elements of the training pipeline, we can conclude regarding their impact on the final model bias. Seven different public BERT models in nine training conditions, i.e. 63 models in total, are compared. Almost all conditions yield significant gender biases. %
Results indicate that reflected biases stem from public BERT models rather than task-specific data, emphasising the weight of responsible usage.  %
\end{abstract}

\begin{figure}[t]
\centering
\includegraphics[width=0.96\linewidth]{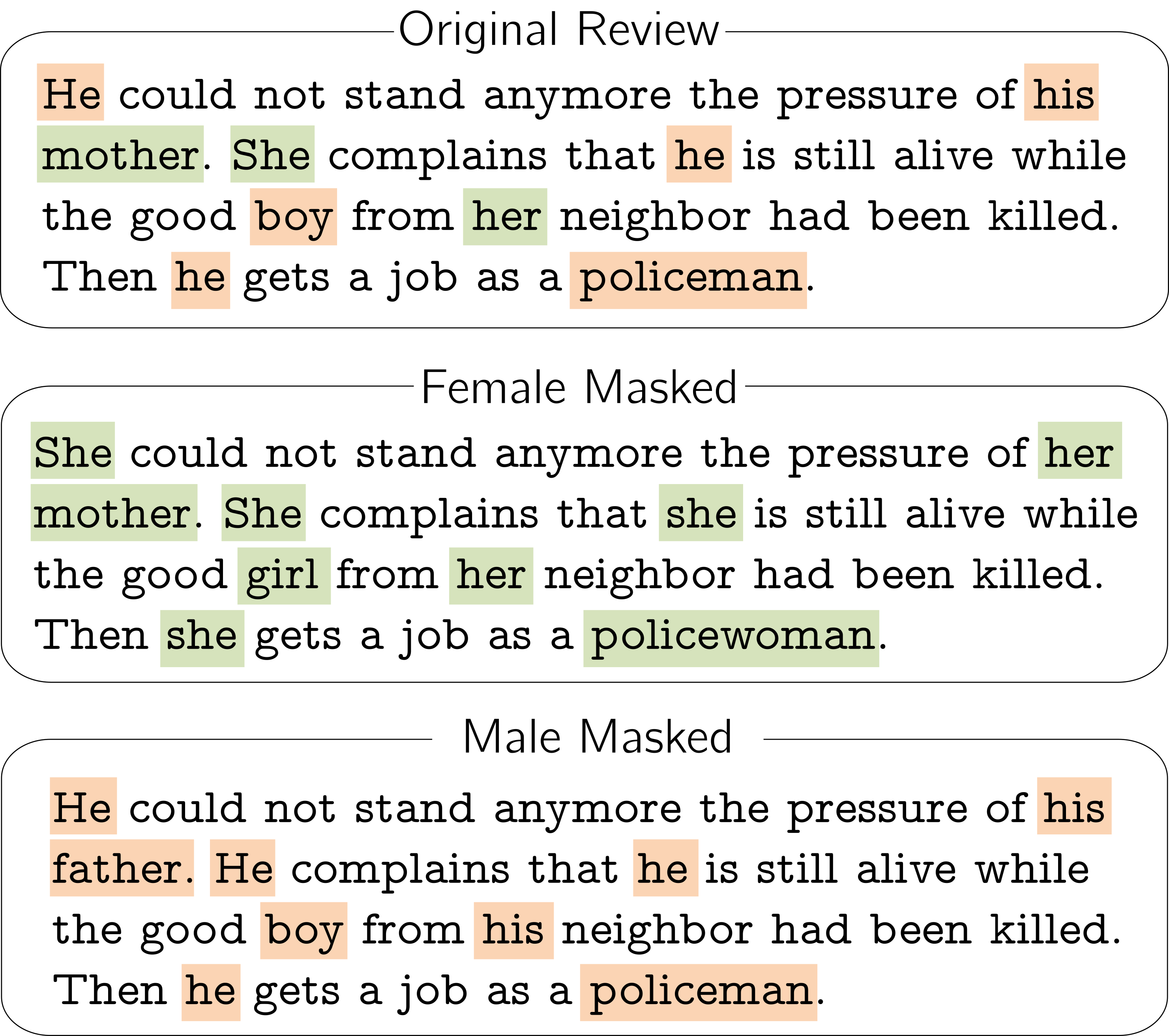}
\caption{\textbf{Example Sample Masking}. The original review contains both male (orange) and female (green) terms. In masked versions, all terms are homogeneously male or female.}
\label{Fig:Example_masking}
\end{figure}

\begin{figure*}[t]
\centering
\includegraphics[width=\textwidth]{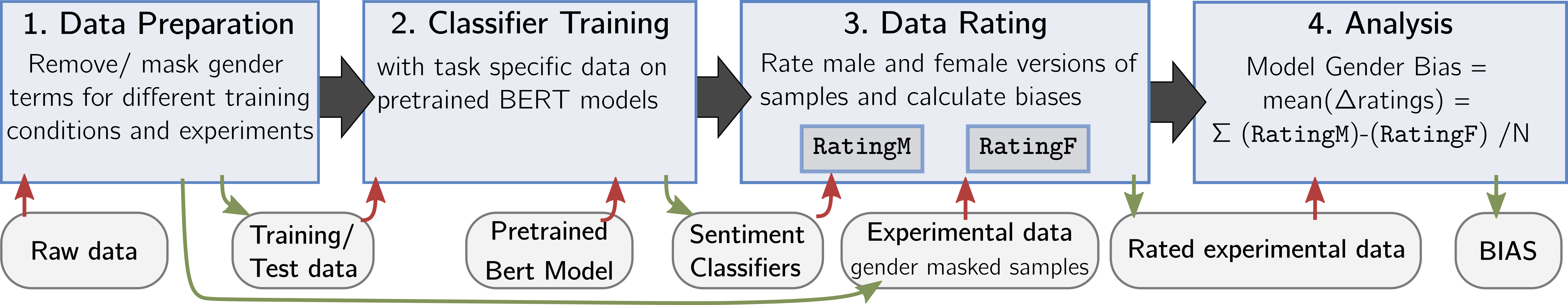} 
\caption{Illustration of experimental pipeline. %
(1) \textbf{Data Preparation}: removing/ balancing gender terms in training data. 
Create experimental data by masking terms in test data to generate a male and a female version. %
(2) \textbf{Model Finetuning}: finetune a pretrained BERT model with task-specific training data in different conditions. 
(3) \textbf{Sentiment Rating} of experimental data; %
(4) \textbf{Analysis} and Bias calculation. \label{Fig:Training_steps}}
\end{figure*}

\section{Introduction}
As complex Machine Learning (ML) based systems are nowadays naturally intertwined with media, technology and everyday life, it is increasingly important to understand their nature and be aware of unwanted behaviour. %
This also applies to the Natural Language Processing (NLP) community, where several recent breakthroughs promoted the application of sophisticated data-driven models in various tasks and applications. %
Only a decade ago, ML-based vector space word embeddings as word2vec \citep{mikolovefficient} or Glove \citep{pennington2014glove} emerged and opened up new ways to extract information and correlations from large amounts of text data. In this context, it has widely been shown that embeddings tend to reflect human biases and stereotypes \citep{caliskan2017semantics,jentzsch2019semantics} and that unintended imbalances in text-embeddings can lead to misbehaviour of systems \citep{bolukbasi2016man}. \\%
In recent years, however, these static word embeddings have rapidly been superseded by the next generation of even more powerful NLP models, which are transformer-based contextualised language models (LMs). BERT \citep{devlin2019bert} and similar architectures established a new standard and now form the basis for many real-life applications and downstream tasks. 
Unfortunately, previous bias measurement approaches do not seem to be straightforwardly transferable \citep{may2019measuring, guo2021detecting, bao2019transfer}. Since the connection between input data and model output is even more opaque, new measures are required to quantify encoded biases in LMs properly. \\
Moreover, with the increase in complexity, computational costs and required amount of data, it is often infeasible to train models from scratch. Instead, pretrained models can be adapted to a wide variety of downstream tasks by finetuning them with a small amount of task-specific data \citep{qiu2020pre}. Although enabling easy access to state-of-the-art NLP techniques, it comes with the risk of lacking model diversity. Different models come with individual characteristics and limitations \citep{xia2020bert}, and there is a small number of well-trained publicly available models that are extensively used, often without even scrutinising the model. %
There are ongoing endeavours to enhance a responsible development and application of those models, e.g. \cite{mitchell2019model}. However, it is still barely understood to what extent biases are propagated through ML pipelines and which training factors enhance or counteract the adaption of discriminating concepts. To apply complex transformer LMs reasonably, it is important to understand how much bias they encode and how these are reflected in downstream applications. \vspace{8pt} \\
%
%
The present study presents a comprehensive analysis of gender bias in BERT models in a downstream sentiment classification task with IMDB data. This task is a realistic scenario, as ML-based recommendation systems are widely used, and reflected stereotypes could directly harm people, e.g. by underrating certain movies or impairing their visibility. %
The investigation comprises two main contributions: First, we propose a novel method to calculate biases in sentiment classification tasks. Sentiment classifiers inherently possess valuation abilities. We exploited these to rate ``female'' and ``male'' sample versions (see Fig.~\ref{Fig:Example_masking}) and therefore need no additional association dimension, e.g. occupation. The classifier is biased if one gender is preferred over the other. %
Second, we analyse the impact of different training factors on the final classifier bias to better understand the origin of biases in NLP tasks. Seven different base models, three different training data conditions and three different bias implementations lead to a total number of 63 compared classifiers. Additional observations could be made regarding training hyperparameters and model accuracies. %
Results reveal significant gender biases in almost all experimental conditions. Compensating imbalances of gender terms in finetuning training data did not show any considerable effect. The size and architecture of pretrained models, in contrast, correlate with the biases. These observations indicate that classifier biases are more likely to stem from the public BERT models than from task-specific data and emphasise the importance of selecting trustworthy resources. %
The present work contributes to understanding biases in language models and how they propagate through complex machine learning systems.

\subsubsection*{Bias Statement} \label{SS:biasStatement}
We study how representational male and female gender concepts are assessed differently in sentiment classification systems. %
In this concrete context, we consider it harmful if a classifier that is trained to distinguish positive and negative movie reviews prefers performers and film characters of one gender over another. This could not only reinforce existing imbalance in the film industry but also lead to direct financial and social harm, e.g. if a movie is less frequently recommended by an automatic recommendation system. \\%
Beyond that, this concrete task is meant to be only one example case for an unlimited number of finetuning scenarios. If we can measure a bias here, this representational imbalance could similarly float into other downstream applications of all kinds, e.g. recruitment processes, hate speech crime detection, news crawler, or computational assistants. %
Generally spoken, it is problematic when freely available and rapidly used models encode a general preference for one gender over another. This is especially critical if this imbalance is propagated through larger systems and unknowingly reflected in gender-unrelated downstream tasks. To raise awareness and mitigate stereotypical reflection, we need to understand how biases emerge and how they are reinforced. \\
The concepts \textit{female} and \textit{male} are represented by sets of terms that are grammatically connected to that gender. One major limitation of that implementation is that it assumes a binary gender classification and does not reflect real-world diversity. Up to now, concepts of a gender-neutral or gender-diverse language are not sufficiently established to consider them for data-driven model training. Nevertheless, we believe that the binary reflection of gender in natural language is worth analysing as it is already connected to real-life discrimination. %

\section{Methodology} \label{S:Introduction}
This investigation analyses to what extent BERT gender biases are present in an IMDB sentiment classification task. We aim to observe what portion of bias emerges in which experimental step by systematically varying conditions in each step. 63 different classifiers and their biases are finally reported in this paper. Many more were trained to observe different training aspects. This section provides a detailed description of experimental steps and how different conditions are achieved. \\%
The experimental pipeline can be divided into four major steps, as illustrated in Fig.~\ref{Fig:Training_steps}. The structure of this section roughly follows these steps. %
First, the preparation of training data is described in Sec.~\ref{ss:data_models}. We compare seven training conditions where gender information in training data is removed or balanced. By that means, it can be measured how much bias is induced during the task-specific finetuning. %
Second, the sentiment classifiers were trained by finetuning seven different common BERT models, as can be read in Sec.~\ref{ss:classifier_training}. By observing whether the choice of model affects the bias magnitude, we can infer how much bias stems from the pretrained BERT model. Also, we compare different sizes of the same architecture. %
In the third step, the trained classifiers were applied to rate the manipulated test data, which is here referred to as experimental data. The setup is described in Sec.~\ref{ss:masking}. Finally, these ratings are used to calculate the model bias that is defined contextually in Sec.~\ref{SS:biasStatement} and mathematically in Sec.~\ref{ss:bias_measure}. Three different sets of gender terms were considered in the experiments. 

\subsection{Sentiment Data and Data Preparation} \label{ss:data_models}
Experiments were conducted in a typical sentiment classification task on movie reviews. %
The \textit{Internet Movie Database}, which is generally referred to as IMDB, is a free platform to rate movies, TV-series and more. We used the publicly available IMDB Large Movie Review Dataset \citep{maas2011learning}, which consists of 50,000 real user movie reviews from that platform. Each sample is provided with the original review texts, the awarded stars as numerical values, and a binary sentiment label derived from the star rating. Reviews with ratings of 4 or lower are labelled as \textit{negative}, and those rated as 7 or higher are labelled as \textit{positive}. Reviews with star ratings of 5 and 6 are not added to the labelled set. The data is already split equally in training and test data, which was not modified in this investigation. We prepared all samples to be free from punctuation and lower-case. The test data was used for model evaluation and also used to create the experimental data as described in Sec.~\ref{ss:masking}. \\
First, each model was trained on the cleaned but unmodified data. This condition is referred to as \textit{original} condition. To see if the occurrence of gender terms in the training data has any effect on the final model biases, we created further conditions. %
Defined gender terms, which are used for bias definition, were fully removed from the training data. This conditions are referred to as \textit{removed}, or specifically \textit{R-pro}, \textit{R-weat}, and \textit{R-all}, for the three different sets of gender terms (see Section~\ref{ss:masking}). %
While removing gender terms is a straightforward step to eliminate them during training, it might lead to incomplete sentences. To see if that affects the results, we defined a third category of training data using Counterfactual Data Augmentation \cite{lu2020gender}. In that approach, a male and a female version of each sample were created by replacing all occurring gender terms (similar to Fig.~\ref{Fig:Example_masking}). Both version are included in the \textit{mixed} training data. This way, each review's structure and completeness are maintained, but the distribution of male and female terms is perfectly balanced. These training conditions are hereafter referred to as \textit{mix-pro}, \textit{mix-weat}, and \textit{mix-all}, respectively. \\
We are aware that neither removing nor mixing gender terms is a mature debiasing technique, as the reflection of gender constructs is much deeper embedded in the language and the content of the text. However, gender bias is here operationalised through different word sets, and by removing those words from training, we aim to avoid changing the learnt associations of the BERT model. %
By that means, we expect to learn whether the positive or negative valuation that is connected to these words stems from the finetuning classifier training or from the previous training of the BERT. 

\begin{figure}[t]
\centering
\includegraphics[width=0.5\textwidth]{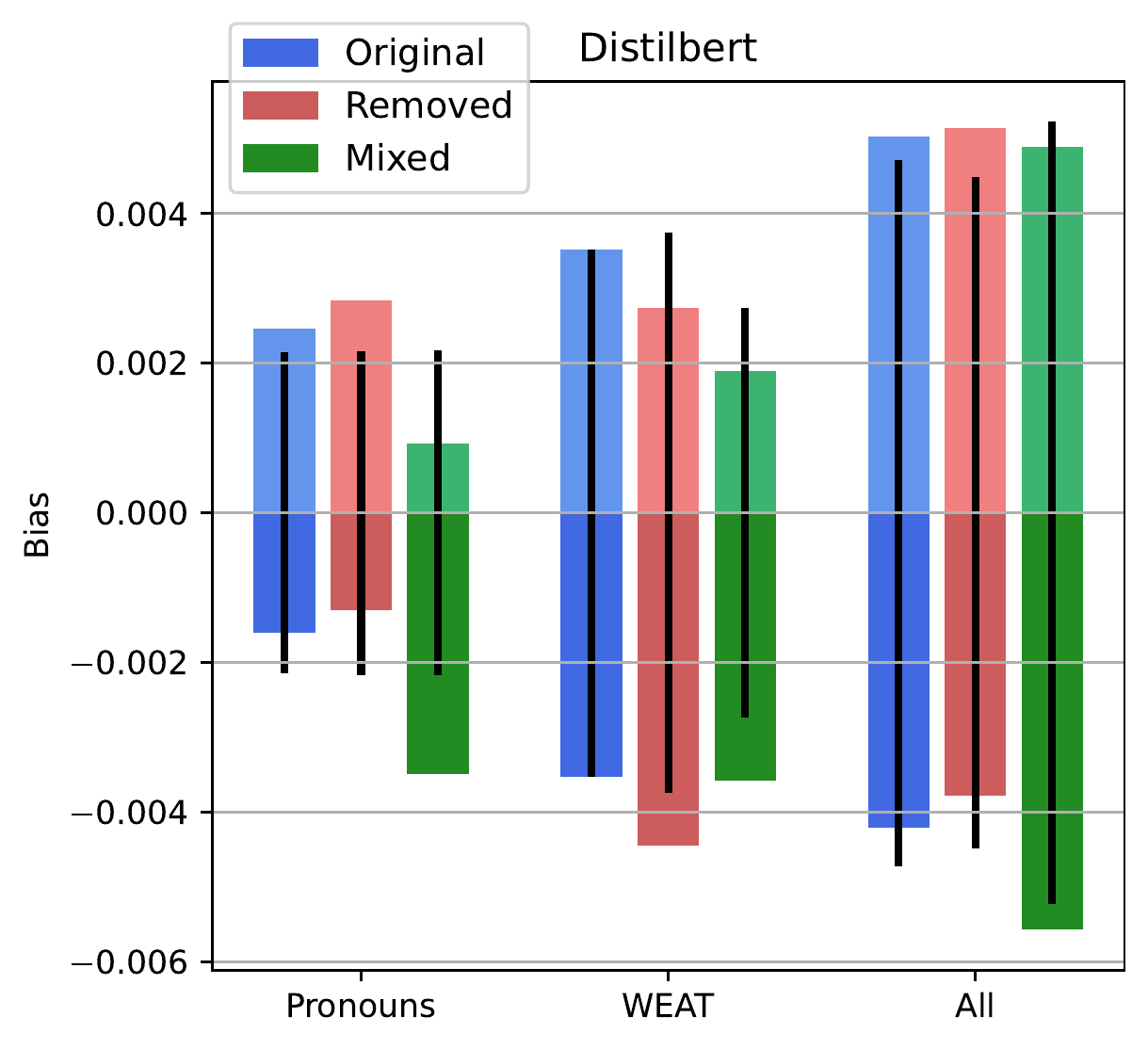} 
\caption{Positive and negative biases for dilstilBERT based classifiers. Blue: trained on original data (\textit{orig.}); red: trained with removed gender terms (\textit{R}); green: trained with mixed gender terms (\textit{mix}). The x-axis is grouped by applied term set (either Pronouns, WEAT, or All). Black lines show the mean total bias symmetrically in both directions to provide an orientation mark for the balance of positive and negative biases.}
\label{Fig:bias_dist}
\end{figure}

\subsection{Classifier Training}\label{ss:classifier_training}
Another main variation between experimental conditions is the selection of a pretrained BERT model. Each classifier is trained by finetuning a pretrained, publicly available BERT model. %
Seven different BERT-based models that differ in architecture and size were selected to examine the effect of model choice on the final bias. The models were provided by HuggingFace \footnote{HuggingFace models, accessed: April 2022. Available at: \url{https://huggingface.co/models}.} and accessed via Transformers Python package \citep{wolf-etal-2020-transformers}. %
All models are trained in a self-supervised fashion without human labelling and on similar training data, which is the Bookscorpus \citep{Zhu_2015_ICCV} and the Englisch Wikipedia\footnote{Wikimedia Foundation, Wikimedia Downloads. Available at: \url{https://dumps.wikimedia.org}}. The following models are considered models in the present analysis:\vspace{12pt}\\
\textbf{DistilBERT} (\textit{distbase}): A smaller and faster version of BERT, 6 layers, 3072 hidden, 12 heads, 66M parameters, vocabulary size: 30522, uncased \citep{Sanh2019DistilBERTAD}.\\
\textbf{BERT base} (\textit{bertbase}): 12 layers, 768 hidden, 12 heads, 110M parameters, vocabulary size: 30522, uncased \citep{DBLP:journals/corr/abs-1810-04805}.\\
\textbf{BERT large} (\textit{bertlarge}): 24 layers, 1024 hidden, 16 heads, 340M parameters, vocabulary size: 30522, uncased \citep{DBLP:journals/corr/abs-1810-04805}.\\
\textbf{RoBERTa base} (\textit{robertbase}): 12 layers, 768 hidden, 12 heads, 125M parameters, vocabulary size: 50265, case-sensitive \citep{DBLP:journals/corr/abs-1907-11692}.\\
\textbf{RoBERTa large} (\textit{robertlarge}): 24 layers, 1024 hidden, 16 heads, 355M parameters, vocabulary size: 50265, case-sensitive \citep{DBLP:journals/corr/abs-1907-11692}.\\
\textbf{AlBERT base} (\textit{albertbase}): 12 layers, 768 hidden, 12 heads, 11M parameters, vocabulary size: 30000, uncased \citep{lan2019albert}. \\
\textbf{AlBERT large} (\textit{albertlarge}): 24 layers, 1024 hidden, 16 heads, 17M parameters, vocabulary size: 30000, uncased \citep{lan2019albert}.\vspace{8pt}\\
The models were trained with a Pytorch framework \citep{NEURIPS2019_9015} on an NVIDIA Tesla V100-SXM3-32GB-H. %
Hyperparameters were inspired by previous literature and kept as constant as possible. However, factors such as different model architectures or the doubled amount of training data in the \textit{mix} conditions required slight adaptions. %
We used a dropout rate of 0.5, which proved to work well in avoiding overfitting. %
Batch sizes were set to be as large as possible, either $32$ or $16$ depending on the model size. The correlation between model accuracy, biases and training batch size was examined and is also elucidated in the results section. %
Learning rates were set between $2e-5$ and $5e-6$ and optimised with Adam. %
As already observed by \citet{de2019understanding}, BERT finetuning tends to overfit quickly. Therefore the authors suggest training for only $2$ to $4$ epochs. Due to extensive hyperparameter optimisation, the present classifiers were trained by finetuning the pretrained models in up to 20 epochs without overfitting. A comprehensive list of all test accuracies and F1 scores can be found in the Appendix. %
The source code of data preparation, model training and experimental analysis will be publicly available on GitHub\footnote{\url{https://github.com/sciphie/bias-bert}}.

\begin{figure*}[t]
\centering
\includegraphics[width=0.45\textwidth]{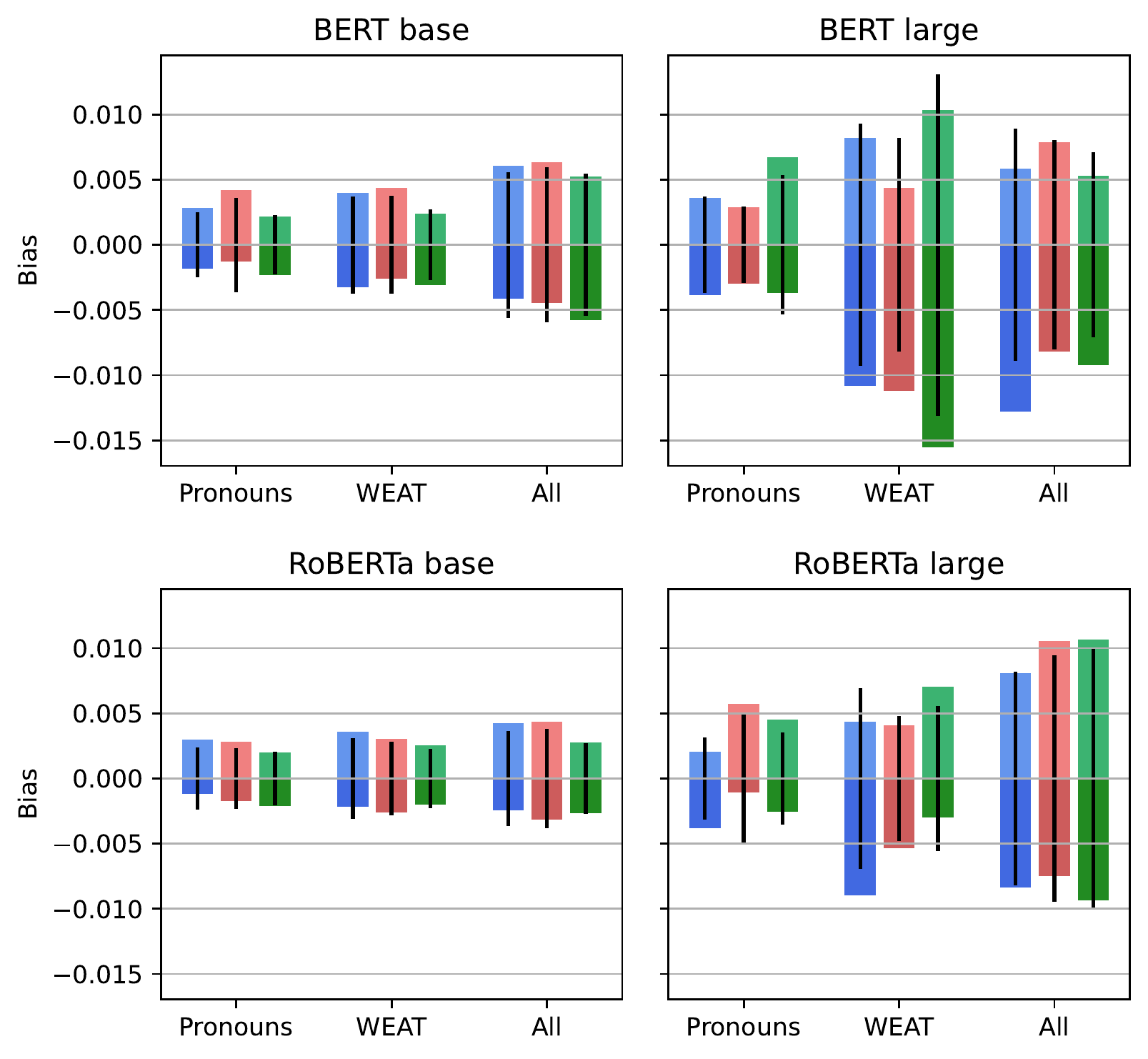}
\includegraphics[width=0.45\textwidth]{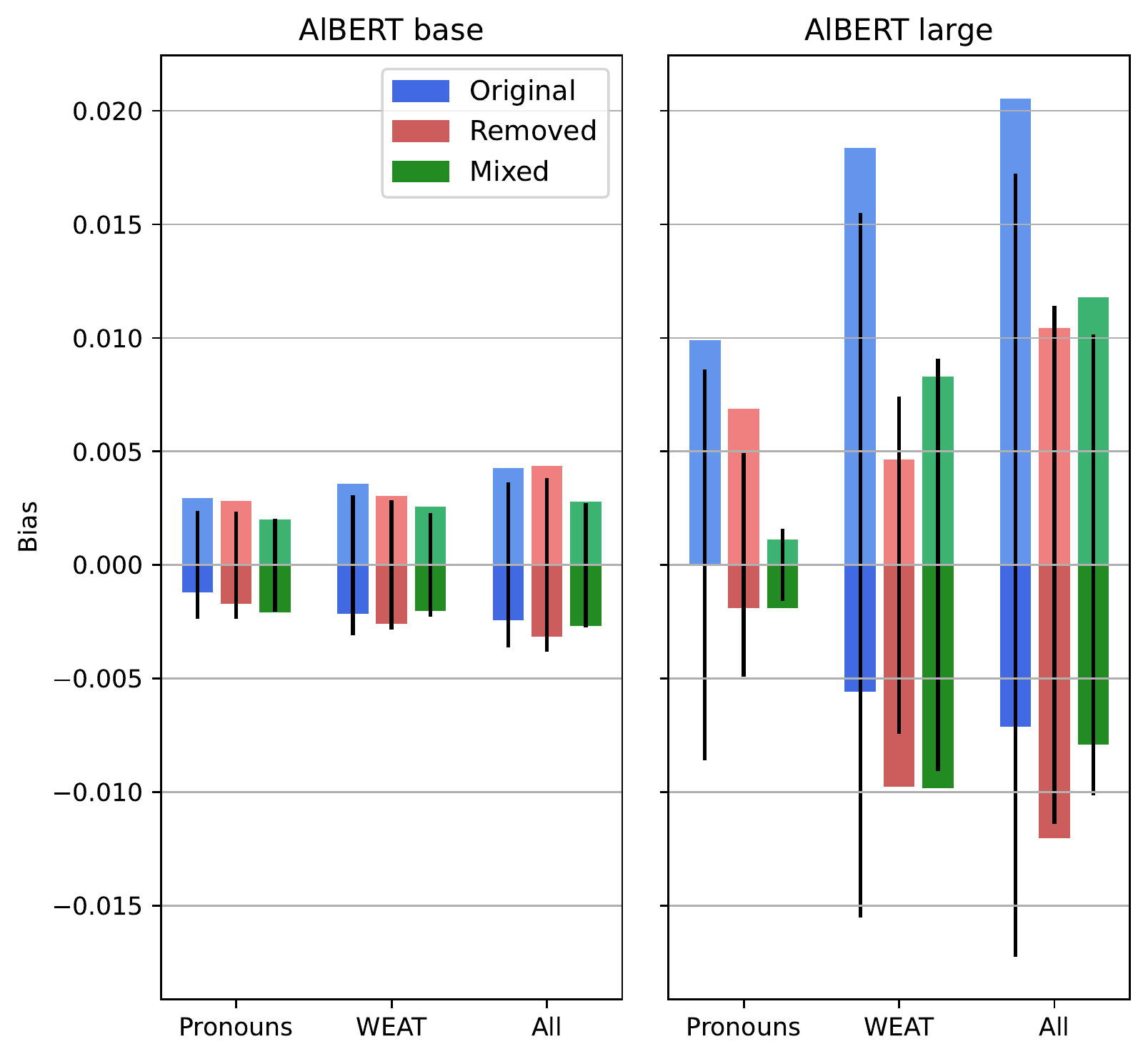}
\caption{Positive and negative biases for classifiers based on different BERT models. Blue: trained on original data (\textit{orig.}); red: trained with removed gender terms (\textit{R}); green: trained with mixed gender terms (\textit{mix}). The x-axis is grouped by applied term set (either Pronouns, WEAT, or All). Black lines show the mean total bias symmetrically in both directions to provide an orientation mark for the balance of positive and negative biases. }
\label{Fig:bias_all}
\end{figure*}

\subsection{Data Masking in Experimental Data}\label{ss:masking}
The analysed bias dimension in this work is \textit{the person being spoken about} \citep{dinan2020multi}, in contrast to, e.g., \citet{excell2021towards} where the bias concerns the author of a comment.
We generated a male ($M$) and a female version ($F$) of each review by turning all included gender terms into the male or female version of that term, respectively. Thus, regardless of whether the terms in the original review were male, female or mixed, the gender of all target terms in each review is homogeneous afterwards (see Fig.~\ref{Fig:Example_masking}). Gender terms were defined in fixed pairs, and only words that occur in the list were masked by their counterpart.\\
The concept of defining and analysing complex construct as the sum of related target and association terms stems originally from the field of psychology \citep{greenwald1998measuring}. This approach has frequently been adapted to computer science and NLP already in the form of the \textit{Word Embedding Association Test} (WEAT) \citep{caliskan2017semantics, jentzsch2019semantics} or similar tasks. \\
The measured bias and the observations in this investigation are likely to depend on the implementation of these target sets to a large extent. Even though many studies apply that approach, the selection of terms is not discussed much. To this end, we created three different sets of target terms to examine the influence of different bias definitions.  %
The largest set comprises all collected gender terms, which is a total number of 341 pairs. In this set, we aimed to collect as many evident gender-specific words as possible. It is named \textit{all} hereafter. A detailed description of the construction of the term set and a list of included words can be found in the Appendix. %
In literature (e.g. WEAT), term lists are usually more compact and restricted to family relations. The second target set is inspired by those resources and consists of 17 word pairs. It is a subset of \textit{all}. We refer to this set as \textit{weat}. The third and smallest set, hereafter named \textit{pro}, only covers pronouns, which are five pairs of terms. This term set is included as pronouns often play a special role in bias research, e.g., in coreference resolution \cite{zhao2018gender}. We seek to understand if pronouns are an adequate bias measure compared to nouns.

\subsection{Bias Measure}\label{ss:bias_measure}
The model bias of a sentiment classifier is determined as follows: %
Two opposite conditions of the bias concept, $X$ and $Y$, are defined and represented by a set of target words, as explained in Sec.~\ref{ss:masking}. For gender bias, these conditions are female $X=F$ and male $Y=M$. %
Test samples, which are a set of natural user comments, are then modified with respect to the bias construct. All naturally included target terms, regardless if they belong to $X$ or $Y$, are replaced by the corresponding terms of either $X$ or $Y$. A male and a female version of each sample are created by that means. The bias for a sample $i$ with X version $i_X$ and Y version $i_Y$ is defined to be the difference between sentiment ratings $sent(i)$ of each version: 
\begin{align}
    Bias_{XY}(i) &= \Delta sent \\
                & = sent(i_Y) - sent(i_X). 
\end{align}
The overall model bias for the sentiment classification system $SC$ is defined to be the mean bias of all $N$ experimental samples: 
\begin{align}
    Bias_{XY}(SC) = \sum_{i=0}^{N} \frac{\Delta sent}{N}  
    \label{for:total_Bias}
\end{align}
As the data classification is binary, the sentiment prediction $sent(i)$ is a scalar value between $0$ and $1$, where $0$ represents the most negative and $1$ the most positive sentiment. Consequently, the sample bias is in the range of $-1$ and $1$, where a high bias value can be interpreted as a bias towards $Y$, i.e. $Y$ is closer associated with positive sentiments than $X$. Analogously, a lower bias value indicates a bias towards $X$, i.e. $X$ being closer associated with positive sentiments. Here, with conditions $M$ and $F$, the total model bias $Bias_{MF} \rightarrow 1$ would indicate a preference for male samples over female ones and $Bias_{FM} \rightarrow -1$ accordingly the other way round. %
Besides the total model bias in Eq. \ref{for:total_Bias}, we also consider the absolute model bias, which is defined as the mean of all absolute biases: 
\begin{align}
    AbsBias_{XY}(SC) =\sum_{i=0}^{N} \frac{|\Delta sent|}{N} 
\end{align}
Analogously, biases will hereafter be referred to as total bias or absolute bias. While the total bias is capable of reflecting the direction of bias, it entails the drawback that contrary sample biases cancel out each other. Therefore the values of absolute biases are stated additionally and quantify the magnitude of bias in the model. \\
We formulated the null and alternative hypotheses for statistical hypothesis testing. Given sample groups $X$ and $Y$ with the medians $m_X$ and $m_Y$  

\begin{description}
    \item[$H_0:$] $m_X = m_Y$: medians are \textbf{equal}; The model is not biased
    \item[$H_A:$] $m_X \neq m_Y$: medians are \textbf{not equal}; The model is considered to be biased
\end{description}
As there are two paired sample groups, which cannot be assumed to be normally distributed, statistics were determined with the Wilcoxon Signed-Rank test. This test has already been applied in similar investigations before, e.g. by \citet{guo2021detecting}). Significance levels are defined as $p<0.05$, $p<0.01$ and $p<0.001$ and are hereafter indicated by one, two and three starlets, respectively. Significance levels were corrected for multiple testing by means of the Bonferroni correction  %
The sample standard deviation normalised by $N-1$ is given by $std$. %
We also state the number of samples below zero, equal to zero and greater than zero to indicate effect sizes. 

\begin{table*}[t]
    \centering
\begin{tabular}{lr | ccc | ccc | ccc }
\hline
       & &  \multicolumn{3}{c | }{pro} &  \multicolumn{3}{c | }{weat}  &  \multicolumn{3}{c }{all}  \\
       \multicolumn{2}{c|}{condition} &  orig. &   R &  mix &  orig. &  R &  mix&  orig. &   R &  mix \\

\hline
\rule{0pt}{3ex}   
 dist   &  abs &    .0021 &  .0022 &   .0022 &  .0035 &   .0037 &  .0027 &      .0047 &   .0045 &    (.0052) \\
        &  tot &    .0009 &  .0010 &  -.0012 &  .0004 &  -.0015 & -.0008 &      .0016 &   .0008 &   (-.0003) \\
\rule{0pt}{3ex}   
 bert B &  abs &    .0025 &  .0036 &   (.0023) &  .0037 &   .0038 &  .0027 &      .0056 &    .0060 &    .0055 \\
        &  tot &    .0013 &  .0031 &  (-.0000) &  .0015 &   .0020 & -.0002 &      .0035 &   .0041 &    .0005 \\
\rule{0pt}{3ex}   
 bert L &  abs &    .0031 &  .0050 &   .0035 &  .0069 &   .0048 &   .0056 &     .0082 &   .0095 &    .0101 \\
        &  tot &   -.0016 &  .0046 &   .0011 & -.0032 &  -.0011 &   .0034 &     .0009 &   .0042 &    .0015 \\
\rule{0pt}{3ex}   
 rob B  &  abs &    .0024 &  .0024 &   .0021 &  .0031 &   .0028 &   (.0023) &     .0036 &   .0038 &    (.0027) \\
        &  tot &    .0016 &  .0009 &  -.0002 &  .0016 &   .0007 &   (.0002) &     .0020 &    .0010 &    (.0000) \\
\rule{0pt}{3ex}   
 rob L  &  abs &    .0024 &  .0025 &   .0020 &  .0039 &   .0039 &   .0028 &     .0044 &   .0043 &    .0041 \\
        &  tot &    .0015 &  .0015 &   .0004 &  .0025 &   .0023 &   .0004 &     .0023 &   .0021 &    .0018 \\
\rule{0pt}{3ex}   
 alb B  &  abs &    .0037 &  .0029 &   .0054 &  .0093 &   .0082 &   .0131 &     .0089 &   .0080 &    .0071 \\
        &  tot &    .0011 & -.0004 &   .0021 &  .0002 &  -.0044 &  -.0034 &    -.0023 &   .0009 &   -.0014 \\
\rule{0pt}{3ex}   
 alb L  &  abs &    .0086 &  .0049 &   .0016 &  .0155 &   .0074 &    (.0091) &    .0172 &   .0114 &    .0101 \\
        &  tot &    .0086 &  .0034 &  -.0008 &  .0130 &  -.0032 &   (-.0009) &    .0137 &  -.0032 &    .0034 \\
\hline
\end{tabular}
\caption{Absolute (abs) and total (tot) model biases of all main experimental classifiers. Positive values indicate a model preference for male samples over female, negative values a preference for female samples over male ones. All biases, except the ones in brackets, are highly significant. Significance levels were Bonferroni corrected for multipe testing. %
\textit{pro}, \textit{weat} and \textit{all} specify the applied term set for training data preprocessing and bias calculation. Terms in training data were either removed (R), balanced (mix), or neither of both (orig.). %
Columns are the different pretrained BERT models used for classifier training. Base models are abbreviated as dist (\textit{distbase}), bert B (\textit{bertbase}), bert L (\textit{bertlarge}), rob B (\textit{robertabase}), rob L (\textit{robertalarge}), alb B (\textit{albertbase}), and alb L (albertlarge).}
    \label{tab:biases_main}
\end{table*}
%
%
\section{Results}
A condensed list of absolute and total biases is reported in Tab.~\ref{tab:biases_main}. Out of 63 reported experimental models, 57 showed highly significant biases. Exceptions are \textit{distbase mix-all}, \textit{bertbase mix-pro}, \textit{robertbase mix-weat}, \textit{robertbase mix-all}, and \textit{albertlarge mix-weat}. 16 classifiers prefer female terms over male terms, and 41 prefer male terms over female terms. Thus, even though more classifiers are in our definition discriminating against women than against men, biases are directed differently. %
The sizes and especially the directions of biases are visualised in Fig.~\ref{Fig:bias_dist} for distilBERT classifiers, in Fig.~\ref{Fig:bias_all} for all other architectures. 

\paragraph{Is bias induced by model finetuning? }
In finetuning systems like this, biases in models can have different origins. We aimed to analyse how much bias was introduced during further training by the task-specific data. To this end, we removed (\textit{R}) or balanced (\textit{mix}) gender terms in the task-specific data to reduce the modification of their representations by finetuning. Both conditions are represented in Fig.~\ref{Fig:bias_dist} and Fig.~\ref{Fig:bias_all} by red and green bars, respectively.\\
Although biases are decreasing by removing gender information from IMDB data in some cases, e.g. \textit{albertalarge pro }, there are likewise examples where it seems to have the opposite effect, such as \textit{bertlarge weat mix}. %
However, for most conditions, these preprocessing measures do not change the magnitude of biases considerably. Especially, removing the gender terms from data does not significantly affect the biases. %
For some models, although the behaviour of the \textit{mix} conditions is different from the other settings, there is no clear pattern observable. Observed differences in that category might also be related to the doubled size of training sets (N = 50000), which is likely to reinforce effects. 

\paragraph{Is bias induced by pretrained models? }
We applied models with different architectures and sizes to observe how measured biases depend on the underlying pretrained model. %
We compare three different sizes of BERT models, which are \textit{distbase}, \textit{bertbase} and \textit{bertlarge}. Moreover, we consider models with RoBERTa architecture in the sizes  \textit{robertabase} and \textit{robertalarge} and AlBERT in \textit{albertbase} and \textit{albertlarge}. 
This comparison leads to two major observations: %
First, biases differ steadily \textit{between} considered architectures. As can be well observed in Fig.~\ref{Fig:bias_dist} and Fig.~\ref{Fig:bias_all}, DistilBERt's biases are about half as big as BERT's and RoBERTa's biases. AlBERTa's biases, again, are about twice as big as those of BERT and RoBERTa. This observation does not only hold among all training conditions but also for both base and large variants. Thus, the architecture of a selected model has an essential impact on the biases of downstream systems. \\
Second, we observe increasing biases depending on model sizes \textit{within} one architecture. \textit{distbase} again yielded the smallest biases, followed by \textit{bertbase}, and \textit{bertlarge}. Simultaneously, \textit{robertalarge} yielded much bigger biases than \textit{robertabase}, and \textit{albertalarge} yielded much bigger biases than \textit{albertbase}. Thus, we observe a correlation between bias and model size, i.e. the number of layers. This indicates that larger models tend to encode greater gender biases. 

\begin{table*}[]
    \centering
\begin{tabular}{l llll} \\
\hline 
        {} &  {bias abs}        & {bias tot}        & {accuracy}        &  {f-score (f1)} \\
\hline \\[-1em]
bias abs    &                   &  \phantom{-}0.373 & -0.481{**}        & -0.497{***}      \\
bias tot    & \phantom{-}0.373  &                   &  \phantom{-}0.044 & -0.085           \\
accuracy    & {-0.481{**}}      &  \phantom{-}0.044 &                   &  \phantom{-}0.886{***} \\
f-score (f1)& {-0.497{***}}     & -0.085            &  \phantom{-}0.886{***} &              \\
\hline
\end{tabular}
    \caption{Correlation (Pearsons) of biases and training details of all 63 classifiers. \textbf{bias abs}: absolute bias, \textbf{bias tot}: total bias.  Starlets indicate levels of significance for $p<0.001$, $p<0.01$ and $p<0.05$, which were Bonferroni corrected for multiple testing.} 
    \label{tab:training_details}
\end{table*} 
%
\paragraph{Is bias dependent on applied term sets?}
As mentioned before, we defined three sets of target terms for the implementation of bias, of which the largest comprises more than a three hundred term pairs and the smallest only five. %
Analogously to term set sizes, the absolute biases are the smallest for the \textit{pronoun} set and the largest for the \textit{all} set in almost all conditions. In other words, the more terms are included, the bigger the measured bias. The only exception is \textit{bertlarge R} and some conditions on \textit{albertbase}. 
Despite the differences in bias magnitude, measured values in all categories were similarly significant. Also, the patterns of effects of training data manipulation or base model comparison could similarly be observed in all three bias definitions. %
We conclude from these observations that all types of included vocabulary encode biases, i.e. pronouns, weat-terms and other nouns. The more terms are included, the higher the measured bias. %
For the presented results, the term set of training data manipulating and the term set for bias measure were always the same. For instance, if we applied the \textit{pro} set to measure biases, we also only removed/ balanced terms of \textit{pro} in the training data. %
We also tested whether biases vary when mixing term sets between different experimental steps. However, that did not reveal any considerable effect, as bias values differed marginally.
\paragraph{Do hyperparameter settings affect biases?} 
Due to computational capacity, some larger models needed to be trained with smaller batch sizes. To see if that affects the final biased, we %
performed additional experiments where we only varied the batch size while fixing all other parameters. For 21 different experimental conditions, models were retrained with batch sizes $32$, $16$ and $8$. Naturally, this affected the course of loss and accuracy during training, but only to a limited extent. All settings led to stable classifiers with convenient model accuracy. The biases of all tested classifiers did not show any indication to be different among the training batch sizes. These results reveal that the batch size does not immediately cause the measured correlation. \\ 
Tab.~\ref{tab:training_details} reports correlations between further basic training details and biases to examine whether there are observable connections. The F-score naturally correlates with accuracy, which is the highest value in the table. %
F1 and accuracy yield a medium negative correlation with absolute biases. 
In contrast to absolute biases, total biases barely show significant correlations with training values. 
All considered classifiers showed good performance in the model evaluation. Test accuracies lie between $77\%$ and $84\%$, which is comparable to baseline values. The evaluation details of all classifiers are attached to the Appendix.

\section{Discussion}
We observed highly significant gender biases in almost all tested conditions. Thus, the present results verify the hypothesis that downstream sentiment classification tasks reflect gender biases.  Although most considered classifiers prefer male samples over female ones, this direction is not consistent: About thirty per cent of classifiers prefer female over male samples. The high significance values are likely to be facilitated by the large sample number and do not necessarily correspond to the effect size. 
It might be insightful to analyse the contexts and types of individual samples to understand how these contrary directions occur. The rating of male and female presence likely depends on the scenario, rather than one gender being strictly advantaged. \\ 
We could not observe any effect of removing gender information from task-specific training data. Thus, in the present case, the biases associated with gender terms are most likely not learnt during finetuning. %
In contrast, results showed significant differences in downstream classifier biases depending on the selection of pretrained models. This is true for both the size and the architecture of the model. %
It is reasonable that the pretrained BERT models, which comprise information from a training set much larger than the IMDB set, are more capable of reflecting complex constructs such as gender stereotypes. It is, therefore, all the more important to develop these models carefully and responsibly and to respect risks and limitations in the application. As a small number of provided base models form the basis for a large portion of applications in NLP, it is especially critical to understand included risks and facilitate debiasing. %
Although the results of the present investigation indicate the origin of biases in pretrained BERT models, that does not preclude the risk to generate biases during finetuning. All elements of the development pipeline need to be audited adequately. \\
We showed that all of the compared term sets are generally appropriate to measure gender bias. However, term sets yielded large differences in bias sizes, showing how crucial the experimental setup is for the validity of measured results. The fact that biases increase relative to the number of gender terms strengthens the conclusion that the majority of these terms reflect biases. It also needs to be further investigated whether the sentiment rating of individual gender terms might be affected by other factors than gender. 
Nevertheless, the applied definition of male and female biases is a rudimentary implementation of real-world circumstances. First, there is a large number of facets that possibly encode gender \citep{doughman2021gender}
, e.g. names or topics. Second, gender is much more diverse in reality than this implementation can reflect. Especially since modern language models are contextual, conceptual stereotypes and biases are likely to be deeply encoded in the embeddings. Automatically learnt models likely cover a large variety of latent biases that contemporary research cannot grasp \citep{gonzalez2020type}. 
This investigation underlines the complexity of bias formation in real-life multi-level systems. 
Results verify the existence of gender biases in BERT's downstream sentiment classification tasks.
In order to further analyse how much of the final system bias stems from the pretrained model, similar experiments could be conducted on debiased BERT models. This way, whether the bias can be further reduced could be tested. Another exciting direction might be to examine how the suggested measurement approach could be transferred to non-binary classification tasks. As a next step, we plan to expand the present experiments to further downstream applications. 

\section{Related Work} \label{S:related Work}
Language models as BERT \citep{devlin2019bert} recently became the new standard in a wide variety of different tasks and superseded static embeddings, as Word2Vec \citep{mikolovefficient} or GloVe \citep{pennington2014glove}. %
For these older embeddings, there already is a huge body of empirical research on bias measuring and mitigation \citep{caliskan2017semantics, jentzsch2019semantics, schramowski2020moral, bolukbasi2016man}, which unfortunately seem to be not straightforwardly tailorable to the new setting \citep{may2019measuring, tan2019assessing}. %
However, recent research finds that BERT also encodes unwanted human biases, such as gender bias \citep{bartl2020unmasking, kurita2019measuring, guo2021detecting}. \\
Downstream task analyses mostly consider shortcomings in dialogue-systems \citep{staliunaite2020compositional, dinan2020queens}. In the context of sentiment analysis, \citet{kiritchenko2018examining} introduced a data set that is designed to measure gender-occupation biases. Although the reported results across 219 tested systems are ambiguous, the framework has been frequently applied ever since \citep{bhardwaj2020investigating, gupta2021evaluating}. %
\citep{huangreducing} measure biases in text-generation systems, i.e. GPT. While the general experimental setting is fundamentally different from the present investigation, they apply a similar idea of measuring biases via sentiment classification. %
To the best of our knowledge, we are the first to utilise sentiment classification to learn about the origin of biases in BERT. %
We contribute to a growing body of exploratory literature regarding bias measure \citep{zhao2020logan, munro2020detecting, field2020unsupervised} and bias mitigation \citep{liu2020mitigating} in contextualised language models.

\section{Conclusion} \label{S:Conclusion}
Contextualised language models such as BERT form the backbone of many everyday applications. 
We introduced a novel approach to measuring bias sentiment classification systems and comprehensively analysed the reflection of gender bias in a realistic downstream sentiment classification task. We compared 63 classifier settings, covering multiple pretrained models and different training conditions. All trained classifiers showed highly significant gender biases. Results indicate that biases are rather propagated from underlying pretrained BERT models than learnt in task-specific training. \\%
Pretrained models should not be applied blindly for downstream tasks as they indeed reflect harmful imbalances and stereotypes. Just as gender-neutral language is important to mitigate everyday discrimination holistically, it is critical to avoid encoded biases in automated systems. We hope that the present work contributes to raising awareness of hidden biases and motivates further research on the propagation of unwanted biases through complex systems. To the best of our knowledge, there is no similar work so far that utilises the valuation capacity of sentiment classifiers to measure downstream biases.


\bibliography{refs} 
\bibliographystyle{acl_natbib}


%
%




\begingroup
\renewcommand\thesection{A}
\section{Appendix}
The following sections supplement presented results with further details. Sec.~\ref{A:SS:word_list} provides all included gender terms and their frequency. Sec.~\ref{A:SS:Biases} presents comprehensive tables with measured biases or all experimental conditions. Sec.~\ref{A:SS:acc} states test accuracies and other evaluation parameters of included classifiers. 
 
\subsection{Target Word Sets} \label{A:SS:word_list}

\textbf{Masked Terms} 
The following list presents all gender terms that were, first, removed and masked to create the training conditions \textit{R} and \textit{mix} and, second, masked with an equivalent term of the opposite gender for experimental data. The list was carefully constructed, incorporating previous literature. \citet{bolukbasi2016man} state a comprehensive list of 218 gender-specific words already. We used that as a root and added further terms that we found in the data itself or other sources and that we considered being missing. Our final list comprises 685 terms in total.\\
In general, if possible, terms were masked by their exact equivalent of the other gender, e.g. \textit{man} by \textit{woman}, and similarly \textit{woman} by \textit{man}. Yet, language and the meaning and connotation of words are highly complex and ambiguous. Thus, the list of terms is not clear-cut, and for some terms, it is disputable whether they should be included or not. These are the four main concerns and how we handled each of them: \vspace{1em}  \\
First, some mappings are not definite, i.e. there are multiple options to transfer the term into the opposite gender. One example is \textit{lady}, which could be the female version of \textit{gentleman} or \textit{lord}. In these cases, we either selected the most likely translation or randomly. \\
Second, some terms do not have an appropriate translation like, among others, the term \textit{guy}, or the term does exist in the other gender but is not used (as much), like for the term \textit{feminism}. In these cases, we tried to find any translation that reflects the meaning as accurate as possible, like \textit{gal} for \textit{guy} or applied the rarely used counterpart, e.g. \textit{masculism}. \\
Third, in some cases, there is a female version of the term, but the male version is usually used for all genders. This is, for example, the case for \textit{manageress} or \textit{lesbianism}. These terms exist and are possibly used, but one could still say 'she is a manager' or 'she is gay'. In these cases, we only translated the term in one direction. This is, whenever the term \textit{lesbian} occurs, it is translated into \textit{gay} for the male version, but when the original rating includes the term \textit{gay}, it is not transformed into \textit{lesbian} for the female version.\\ 
Finally, it can have other meanings that are not gender-related, e.g. \textit{Miss} as an appellation can also be the verb \textit{to miss}. We decided to interpret these terms as the more frequent meaning or to leave the term out if it was unclear.\\
Similar to many other resources, \citet{bolukbasi2016man} also include terms from the animal realm, such as \textit{stud} or \textit{lion}\cite{bolukbasi2016man}. We decided not to do so because the present investigation focuses on human gender bias, which might not be similarly present for animals. %
The list includes all masked terms that occurred at least ten times in the entire experimental data in decreasing order. Further 404 terms were included in the analysis that occurred fewer than ten times. 221 of these terms were not counted even once and did not affect the analysis. A comprehensive list of all considered terms and their frequency can be found in the corresponding repository. \\
The full list corresponds to the \textit{all} term set. Due to the above-discussed concerns, we also applied the \textit{weat} term set, which consists of mostly unambiguous terms. Terms that are included in weat are marked in bold. The third term set, \textit{pro}, only includes pronouns which are \textit{he, she, his, her, him} and \textit{hers}. This term set is relatively small, but pronouns are more frequent than most other terms. 

Pronouns are marked in bold.  \vspace{1em}\\
\textbf{he} (46634), 
\textbf{his} (34475), 
\textbf{her} (31303), 
\textbf{she} (26377), 
\textbf{him} (17863), 
\textbf{man} (11656), 
guys (8070), 
\textbf{girl} (7433), 
guy (5862), 
god (5324), 
mom (4456), 
actors (4349), 
\textbf{boy} (3802), 
\textbf{girls} (3509), 
\textbf{mother} (3424), 
dad (3274), 
\textbf{woman} (3235), 
wife (2858), 
\textbf{brother} (2810), 
\textbf{sister} (2726), 
\textbf{men} (2662), 
\textbf{father} (2468), 
mr (2439), 
\textbf{boys} (2377), 
actor (2369), 
\textbf{son} (2226), 
\textbf{women} (2212), 
himself (2194), 
dude (2089), 
\textbf{daughter} (1995), 
lady (1948), 
husband (1658), 
boyfriend (1544), 
\textbf{brothers} (1474), 
hero (1427), 
actress (1167), 
\textbf{female} (1158), 
girlfriend (1087), 
king (1012), 
\textbf{mothers} (1009), 
hubby (994), 
count (932), 
herself (878), 
\textbf{male} (821), 
daddy (792), 
ladies (766), 
ms (725), 
giant (725), 
mommy (721), 
master (708), 
\textbf{sisters} (701), 
lord (697), 
ma (671), 
sir (626), 
queen (621), 
mama (596), 
\textbf{uncle} (587), 
chick (567), 
moms (556), 
grandma (529), 
\textbf{aunt} (521), 
\textbf{fathers} (444), 
heroes (434), 
princess (432), 
pa (411), 
host (405), 
niece (373), 
prince (350), 
dads (341), 
actresses (341), 
priest (328), 
nephew (328), 
hunter (303), 
bride (284), 
witch (281), 
lesbian (277), 
heroine (261), 
kings (239), 
grandpa (239), 
daughters (234), 
\textbf{grandfather} (223), 
\textbf{grandmother} (222), 
chicks (193), 
masters (187), 
cowboy (185), 
counts (177), 
dudes (174), 
sons (169), 
gods (166), 
gal (159), 
papa (158), 
wifey (156), 
girly (156), 
queens (152), 
bachelor (149), 
housewives (148), 
\textbf{hers} (148), 
maid (145), 
girlfriends (145), 
beard (141), 
emperor (136), 
gentleman (129), 
superman (128), 
duke (127), 
girlie (125), 
mayor (123), 
wives (122), 
gentlemen (116), 
playboy (114), 
mister (113), 
mistress (111), 
giants (109), 
females (107), 
wizard (105), 
widow (98), 
nun (98), 
penis (96), 
fiance (95), 
lad (92), 
gals (92), 
boyfriends (91), 
girlies (90), 
bloke (90), 
bachelorette (88), 
aunts (87), 
policeman (84), 
males (84), 
fella (79), 
diva (79), 
macho (78), 
goddess (78), 
lads (77), 
landlord (75), 
fiancé (75), 
patron (74), 
waitress (73), 
husbands (70), 
hosts (70), 
fiancée (70), 
feminist (70), 
cowboys (70), 
nephews (68), 
mermaid (68), 
sorority (66), 
grandmas (66), 
chap (65), 
manly (64), 
businessman (63), 
monk (62), 
baron (62), 
witches (61), 
bachelor (61), 
nieces (59), 
housewife (59), 
\textbf{feminine} (58), 
cameraman (58), 
shepherd (57), 
lesbians (55), 
vagina (53), 
uncles (53), 
wizards (52), 
henchmen (49), 
salesman (48), 
postman (48), 
mamas (48), 
grandson (48), 
brotherhood (47), 
lords (44), 
henchman (44), 
waiter (43), 
dukes (42), 
mommies (41), 
fellas (41), 
granddaughter (40), 
traitor (39), 
groom (39), 
duchess (39), 
madman (36), 
policemen (35), 
conductor (35), 
sisterhood (34), 
fraternity (34), 
monks (33), 
\textbf{masculine} (33), 
nuns (32), 
fiancee (32), 
lass (30), 
tailor (29), 
priests (29), 
maternity (29), 
butch (29), 
stepfather (28), 
hostess (28), 
ancestors (28), 
heiress (27), 
countess (27), 
congressman (27), 
bridesmaid (27), 
protector (26), 
divas (26), 
ambassador (26), 
damsel (25), 
steward (24), 
madam (24), 
homeboy (24), 
landlady (23), 
\textbf{grandmothers} (23), 
fireman (23), 
empress (23), 
chairman (23), 
widower (22), 
sorcerer (22), 
patrons (22), 
masculinity (22), 
firemen (22), 
englishman (22), 
businessmen (22), 
testosterone (21), 
manhood (21), 
chaps (21), 
widows (20), 
lesbianism (20), 
blokes (20), 
beards (20), 
barbershop (20), 
anchorman (20), 
sperm (19), 
heroines (19), 
heir (19), 
stepmother (18), 
princesses (18), 
princes (18), 
handyman (18), 
patriarch (17), 
monastery (17), 
mailman (17), 
homegirl (17), 
headmistress (17), 
fisherman (17), 
czar (17), 
brotherly (17), 
brides (17), 
uterus (16), 
maternal (16), 
abbot (16), 
prophet (15), 
boyish (15), 
adventurer (15), 
testicles (14), 
temptress (14), 
schoolgirl (14), 
penises (14), 
maids (14), 
barmaid (14), 
waiters (13), 
traitors (13), 
stuntman (13), 
priestess (13), 
seductress (12), 
schoolboy (12), 
motherhood (12), 
daddies (12), 
cowgirls (12), 
cameramen (12), 
bachelors (12), 
adventurers (12), 
sculptor (11), 
schoolgirls (11), 
proprietor (11), 
paternal (11), 
homeboys (11), 
foreman (11), 
feminism (11), 
doorman (11), 
bachelors (11), 
womanhood (10), 
testicle (10), 
mistresses (10), 
merman (10), 
\textbf{grandfathers} (10), 
girlish (10)

\subsection{Biases}\label{A:SS:Biases}
Tab.~\ref{A:Tab:bias-overview1} and Tab.~\ref{A:Tab:bias-overview2} provide an overview of the model biases of all considered classifiers. %
For the calculation of biases, the same gender term set was applied to the experimental data masking as for the training data condition. This means, for instance, in the experimental data for all \textit{R-weat} and \textit{mix-weat} trained classifiers, only \textit{weat} terms were masked. Thus, the training condition is in line with the experimental bias calculation for all \textit{N} and \textit{mix} training conditions. For \textit{original} training conditions, however, no term set was applied to the training data. This is why biases of all three term groups are compared, which are \textit{original-N}, \textit{original-pro}, and \textit{original-weat}. \\%
Wilcoxon signed-rank-test yielded highly significant p-values for almost all conditions. Exceptions are \textit{distbert mix-all},\textit{bertbase mix-pro}, \textit{robertbase mix-weat}, \textit{robertbase mix-all}, and \textit{albertlarge mix-weat}. %
Out of 63 reported experimental models, 57 showed highly significant biases, of which 16 prefer female terms over male terms, and 41 prefer male terms over female terms. 

\begin{table*}[p]
\centering
\renewcommand{\arraystretch}{1.2}
\begin{tabular}{l l | c c | c c  |  r c l | c } 
    \hline 
        &           & \multicolumn{2}{c|}{non zero } & \multicolumn{2}{c|}{all}  &  &  &  &   \\
        & Condition & bias abs & bias tot & bias abs & bias tot & N$<0$ & N$=0$ & N$>0$ & sign.  \\
    \hline 
\multicolumn{2}{l | }{distbert }&&&&&&&&\\
    &  original-pro     &  0.0021 &  0.0009 &  0.0014 &  0.0006 &  6085 &  10216 &  8699 & {***} \\ 
    &  R-pro            &  0.0022 &  0.0010 &  0.0014 &  0.0007 &  7116 &  9183 &  8701 & {***} \\  
    &  mix-pro          &  0.0022 &  -0.0012 &  0.0012 &  -0.0007 &  6922 &  7309 &  10769 & {***} \\  
    &  original-weat    &  0.0035 &  0.0004 &  0.0026 &  0.0003 &  8098 &  10214 &  6688 & {***} \\ 
    &  R-weat           &  0.0037 &  -0.0015 &  0.0027 &  -0.0011 &  10773 &  7532 &  6695 & {***} \\  
    &  mix-weat         &  0.0027 &  -0.0008 &  0.0018 &  -0.0006 &  8332 &  8428 &  8240 & {***} \\  
    &  original-all     &  0.0047 &  0.0016 &  0.0039 &  0.0013 &  7817 &  12941 &  4242 & {***} \\ 
    &  R-all            &  0.0045 &  0.0008 &  0.0037 &  0.0007 &  10022 &  10734 &  4244 & {***} \\  
    &  mix-all          &  0.0052 &  -0.0003 &  0.0042 &  -0.0003 &  10080 &  10177 &  4743 & {-} \\ 
\multicolumn{2}{l | }{bertbase }&&&&&&&&\\
    &  original-pro     &  0.0025 &  0.0013 &  0.0016 &  0.0008 &  5430 &  10874 &  8696 & {***} \\ 
    &  R-pro            &  0.0036 &  0.0031 &  0.0024 &  0.0020 &  3234 &  13061 &  8705 & {***} \\  
    &  mix-pro          &  0.0023 & -0.0000 &  0.0014 &  -0.0000 &  7505 &  7794 &  9701 & {-} \\  
    &  original-weat    &  0.0037 &  0.0015 &  0.0027 &  0.0011 &  6187 &  12128 &  6685 & {***} \\ 
    &  R-weat           &  0.0038 &  0.002 &  0.0028 &  0.0015 &  6204 &  12098 &  6698 & {***} \\  
    &  mix-weat         &  0.0027 &  -0.0002 &  0.0015 &  -0.0001 &  6421 &  7135 &  11444 & {***} \\  
    &  original-all     &  0.0056 &  0.0035 &  0.0046 &  0.0029 &  5233 &  15527 &  4240 & {***} \\ 
    &  R-all            &  0.0060 &  0.0041 &  0.0049 &  0.0034 &  4319 &  16431 &  4250 & {***} \\  
    &  mix-all          &  0.0055 &  0.0005 &  0.0035 &  0.0003 &  6838 &  9001 &  9161 & {***} \\  
\multicolumn{2}{l | }{bertlarge }&&&&&&&&\\
    &  original-pro     &  0.0031 &  -0.0016 &  0.0021 &  -0.0011 &  10287 &  6020 &  8693 & {***} \\  
    &  R-pro            &  0.0050 &  0.0046 &  0.0032 &  0.0030 &  2697 &  13610 &  8693 & {***} \\  
    &  mix-pro          &  0.0035 &  0.0011 &  0.0014 &  0.0004 &  4961 &  5228 &  14811 & {*} \\ 
    &  original-weat    &  0.0069 &  -0.0032 &  0.0051 &  -0.0023 &  10329 &  7986 &  6685 & {***} \\  
    &  R-weat           &  0.0048 &  -0.0011 &  0.0035 &  -0.0008 &  10172 &  8142 &  6686 & {***} \\  
    &  mix-weat         &  0.0056 &  0.0034 &  0.0029 &  0.0018 &  4581 &  8195 &  12224 & {***} \\  
    &  original-all     &  0.0082 &  0.0009 &  0.0068 &  0.0007 &  9128 &  11633 &  4239 & {***} \\  
    &  R-all            &  0.0095 &  0.0042 &  0.0079 &  0.0035 &  7314 &  13443 &  4243 & {***} \\  
    &  mix-all          &  0.0101 &  0.0015 &  0.0078 &  0.0012 &  8848 &  10455 &  5697 & {***} \\  
\hline
\end{tabular}
\caption{Total biases of all experimental classifiers (part 1). The bias is the mean bias over all experimental samples. While the absolute bias (bias abs) is the mean of absolute values, the total bias (bias tot) is based on the directed sample biases. For ``non zero" values, samples with a bias$=0$ are excluded. ``all" includes all 25000 sample biases. %
The numbers of samples with negative, no, and positive bias are given by $N<0$, $N=0$, or $N>0$, respectively. %
Significance levels for Wilcoxon signed-rank-test were defined as $p>0.05 : ${*}, $p>0.01 : ${**}, and $p>0.001 : ${***}. Reported significance levels were corrected for multiple testing with the Bonferroni correction. }
\label{A:Tab:bias-overview1}
\end{table*}

\begin{table*}[p]
\centering
\renewcommand{\arraystretch}{1.2}
\begin{tabular}{l l | c c | c c |  r c l  | c } 
    \hline 
        &           & \multicolumn{2}{c|}{non zero } & \multicolumn{2}{c|}{all}  &  &  &  &   \\
        & Condition & bias abs & bias tot & bias abs & bias tot & N$<0$ & N$=0$ & N$>0$ & sign.  \\
\hline 
\multicolumn{2}{l | }{robertabase} &&&&&&&&\\
    &  original-pro &  0.0024 &  0.0016 &  0.0015 &  0.0010 &  5448 &  10840 &  8712 & {***} \\ 
    &  R-pro        &  0.0024  &  0.0009 &  0.0015 &  0.0006 &  6822 &  9472 &  8706    & {***} \\  
    &  mix-pro      &  0.0021 &  -0.0002 &  0.0013 &  -0.0001 &  8682 &  7612 &  8706    & {***} \\  
    &  original-weat&  0.0031 &  0.0016 &  0.0023 &  0.0011 &  6470 &  11832 &  6698    & {***} \\ 
    &  R-weat       &  0.0028 &  0.0007 &  0.0021 &  0.0005 &  7722 &  10581 &  6697    & {***} \\  
    &  mix-weat     &  0.0023 &  0.0002 &  0.0017 &  0.0002 &  9396 &  8894 &  6710    & {-} \\  
    &  original-all &  0.0036 &  0.0020 &  0.0030 &  0.0016 &  7165 &  13585 &  4250    & {***} \\  
    &  R-all        &  0.0038 &  0.0010 &  0.0032 &  0.0008 &  9294 &  11464 &  4242    & {***} \\  
    &  mix-all      &  0.0027 & 0.0000 &  0.0023 &  0.0000 &  10520 &  10206 &  4274    &  {-} \\  
\multicolumn{2}{l | }{robertalarge}&&&&&&&&\\
    &  original-pro &  0.0024 &  0.0015 &  0.0016 &  0.0010 &  5235 &  11055 &  8710    & {***} \\  
    &  R-pro        &  0.0025 &  0.0015 &  0.0016 &  0.0010 &  5216 &  11072 &  8712    & {***} \\  
    &  mix-pro      &  0.0020 &  0.0004 &  0.0013 &  0.0003 &  6679 &  9606 &  8715    & {***} \\  
    &  original-weat&  0.0039 &  0.0025 &  0.0029 &  0.0018 &  5894 &  12411 &  6695    & {***} \\ 
    &  R-weat       &  0.0039 &  0.0023 &  0.0029 &  0.0017 &  6109 &  12193 &  6698    & {***} \\  
    &  mix-weat     &  0.0028 &  0.0004 &  0.0021 &  0.0003 &  8071 &  10220 &  6709    & {***} \\  
    &  original-all &  0.0044 &  0.0023 &  0.0036 &  0.0019 &  7105 &  13653 &  4242    & {***} \\ 
    &  R-all        &  0.0043 &  0.0021 &  0.0035 &  0.0017 &  7045 &  13712 &  4243    & {***} \\  
    &  mix-all      &  0.0041 &  0.0018 &  0.0034 &  0.0015 &  6971 &  13783 &  4246    & {***} \\  
\multicolumn{2}{l | }{albertbase}&&&&&&&&\\
    &  original-pro &  0.0037 &  0.0011 &  0.0024 &  0.0007 &  5481 &  10811 &  8708    & {***} \\ 
    &  R-pro        &  0.0029 &  -0.0004 &  0.0019 &  -0.0003 &  9305 &  6986 &  8709    & {***} \\  
    &  mix-pro      &  0.0054 &  0.0021 &  0.0035 &  0.0014 &  7244 &  8968 &  8788    & {***} \\  
    &  original-weat&  0.0093 &  0.0002 &  0.0068 &  0.0001 &  7710 &  10600 &  6690    & {***} \\ 
    &  R-weat       &  0.0082 &  -0.0044 &  0.006 &  -0.0032 &  10346 &  7942 &  6712    & {***} \\  
    &  mix-weat     &  0.0131 &  -0.0034 &  0.0093 &  -0.0024 &  9426 &  8263 &  7311    & {***} \\  
    &  original-all &  0.0089 &  -0.0023 &  0.0074 &  -0.0019 &  9112 &  11645 &  4243    & {***} \\  
    &  R-all        &  0.0080 &  0.0009 &  0.0067 &  0.0008 &  8979 &  11769 &  4252    & {***} \\  
    &  mix-all      &  0.0071 &  -0.0014 &  0.0058 &  -0.0012 &  9481 &  11030 &  4489    & {***} \\  
\multicolumn{2}{l | }{albertlarge}&&&&&&&&\\
    &  original-pro &  0.0086 &  0.0086 &  0.0056 &  0.0056 &  2120 &  14075 &  8805    & {***} \\ 
    &  R-pro        &  0.0049 &  0.0034 &  0.0032 &  0.0022 &  6407 &  9869 &  8724    & {***} \\  
    &  mix-pro      &  0.0016 &  -0.0008 &  0.0010&  -0.0005 &  10121 &  6136 &  8743    & {***} \\  
    &  original-weat&  0.0155 &  0.0130 &  0.0113 &  0.0095 &  4058 &  14191 &  6751    & {***} \\ 
    &  R-weat       &  0.0074 &  -0.0032 &  0.0054 &  -0.0023 &  9936 &  8373 &  6691    & {***} \\  
    &  mix-weat     &  0.0091 &  -0.0009 &  0.0066 &  -0.0006 &  9186 &  8998 &  6816    & {-} \\  
    &  original-all &  0.0172 &  0.0137 &  0.0143 &  0.0114 &  5095 &  15594 &  4311    & {***} \\ 
    &  R-all        &  0.0114 &  -0.0032 &  0.0095 &  -0.0026 &  12573 &  8180 &  4247    & {***} \\  
    &  mix-all      &  0.0101 &  0.0034 &  0.0084 &  0.0028 &  8777 &  11875 &  4348 & {***} \\  
\hline
\end{tabular}
\caption{Total biases of all experimental classifiers (part 2). Extension of Tab.~\ref{A:Tab:bias-overview1}}
\label{A:Tab:bias-overview2}
\end{table*}

\newpage
\begin{table}[t]
    \centering
\begin{tabular}{lll cccc}
\hline
    \multicolumn{2}{l}{Model / Spec} &  acc. &   rec. &  prec. & f1 \\
    \hline    
    \multicolumn{2}{l}{distbase}  &&&&\\
                &   original &   .812 &  .778 &   .835 &  .805 \\    
                &     R-all &   .817 &  .789 &   .836 &  .812 \\
                &    R-weat &   .820 &  .789 &   .840 &  .814 \\                
                &     R-pro &   .818 &  .780 &   .844 &  .811 \\
                &   mix-all &   .822 &  .795 &   .840 &  .817 \\
                &  mix-weat &   .822 &  .783 &   .849 &  .815 \\                
                &   mix-pro &   .822 &  .784 &   .848 &  .815 \\
   \multicolumn{2}{l}{ bertbase}  &&&&\\
                &   original &   .818 &  .787 &   .838 &  .812 \\   
                &     R-all &   .821 &  .781 &   .849 &  .813 \\
                &     R-pro &   .820 &  .776 &   .851 &  .812 \\
                &    R-weat &   .821 &  .803 &   .833 &  .818 \\
                &   mix-all &   .836 &  .791 &   .868 &  .828 \\
                &   mix-pro &   .835 &  .816 &   .849 &  .832 \\
                &  mix-weat &   .835 &  .812 &   .852 &  .832 \\
    \multicolumn{2}{l}{bertlarge}  &&&&\\
                 &   original &   .805 &  .787 &   .816 &  .801 \\    
                &     R-all &   .797 &  .734 &   .839 &  .783 \\
                &     R-pro &   .779 &  .660 &   .867 &  .749 \\
                &    R-weat &   .803 &  .739 &   .847 &  .789 \\
                &   mix-all &   .795 &  .723 &   .845 &  .780 \\
                &   mix-pro &   .797 &  .738 &   .836 &  .784 \\
                &  mix-weat &   .789 &  .710 &   .843 &  .771 \\
     \hline
    \end{tabular}
    \caption{Test accuracy (acc.), recall (rec.), precision (prec.), and F1-Score (f1) for the models that are used in the experiments - part 1}
    \label{A:tab:acc1}
\end{table}

\subsection{Evaluation of Models} \label{A:SS:acc}
Tab.~\ref{A:tab:acc1} and Tab.~\ref{A:tab:acc2} show the accuracies, recalls, precisions and F1-Score of all experimental models calculated on the test data. For the calculation of reported values, the test data set has been treated analogously to the training condition. That means for instance, since we removed all pronouns from training data in the R-all condition, we did the same in the test data before evaluating the models in that condition. 

\begin{table}[t]
    \centering
\begin{tabular}{lll cccc}
\hline
    \multicolumn{2}{l}{Model / Spec} &  acc. &   rec. &  prec. & f1 \\
    \hline
     \multicolumn{2}{l}{robertabase}  &&&&\\
                &   original &   .818 &  .744 &   .874 &  .804 \\     
                &     R-all &   .823 &  .770 &   .862 &  .813 \\
                &    R-weat &   .820 &  .739 &   .881 &  .804 \\                
                &     R-pro &   .818 &  .733 &   .883 &  .801 \\
                &   mix-all &   .833 &  .780 &   .873 &  .824 \\
                &  mix-weat &   .830 &  .781 &   .867 &  .821\\               
                &   mix-pro &   .823 &  .760 &   .870 &  .811 \\
    \multicolumn{2}{l}{robertalarge}  &&&&\\
                &   original &   .820 &  .748 &   .873 &  .806 \\     
                &    R-all &   .820 &  .765 &   .859 &  .810 \\
                &    R-weat &   .820 &  .761 &   .862 &  .809 \\                
                &    R-pro &   .818 &  .751 &   .868 &  .805 \\
                &   mix-all &   .815 &  .749 &   .862 &  .801 \\
                &  mix-weat &   .816 &  .761 &   .855 &  .805 \\                
                &   mix-pro &   .814 &  .728 &   .879 &  .797 \\
    \multicolumn{2}{l}{albertbase}  &&&&\\
                &  original  &   .693 &  .932 &   .630 &  .752 \\    
                &     R-all &   .771 &  .711 &   .809 &  .756 \\
                &    R-weat &   .772 &  .749 &   .785 &  .767 \\                
                &     R-pro &   .757 &  .748 &   .764 &  .756 \\
                &   mix-all &   .782 &  .791 &   .777 &  .784 \\
                &  mix-weat &   .778 &  .818 &   .757 &  .786 \\                
                &   mix-pro &   .780 &  .813 &   .762 &  .787 \\
    \multicolumn{2}{l}{albertlarge}  &&&&\\
                &   original &   .784 &  .762 &   .797 &  .779 \\    
                &     R-all &   .762 &  .847 &   .724 &  .781 \\
                &    R-weat &   .767 &  .802 &   .750 &  .775 \\                
                &     R-pro &   .763 &  .832 &   .732 &  .779 \\
                &   mix-all &   .774 &  .803 &   .759 &  .781 \\
                &  mix-weat &   .784 &  .788 &   .781 &  .785 \\                
                &   mix-pro &   .782 &  .752 &   .801 &  .776 \\
    \hline
    \end{tabular}
    \caption{Test accuracy (acc.), recall (rec.), precision (prec.), and F1-Score (f1) for the models that are used in the experiments - part 2}
    \label{A:tab:acc2}
\end{table}
\clearpage

\endgroup


\end{document}